\documentclass[lettersize,journal]{IEEEtran}
\usepackage{amsmath,amsfonts}
\usepackage{algorithmic}
\usepackage{algorithm}
\usepackage{array}
\usepackage[caption=false,font=normalsize,labelfont=sf,textfont=sf]{subfig}
\usepackage{textcomp}
\usepackage{stfloats}
\usepackage{url}
\usepackage{verbatim}
\usepackage{graphicx}
\usepackage{cite}
\usepackage{makecell}
\usepackage[accsupp]{axessibility}  % Improves PDF readability for those with disabilities.
\usepackage{multirow} 
\usepackage{tabularx}
\usepackage{xcolor} 
\usepackage{booktabs} 
\usepackage{graphicx}   
\usepackage[table]{xcolor}  
\usepackage{booktabs, graphicx, multirow} 
\usepackage{comment}
\usepackage{amssymb}
\definecolor{mycolor}{HTML}{DCEFF5}
\definecolor{grey1}{rgb}{0.9,0.9,0.9}
\definecolor{white}{rgb}{1,1,1}

\begin{document}

\title{Robust Vision-Language Navigation with Awareness of Instruction-Execution Progress}

\author{Jiagan~Huang,
        Zikun~Zhou,
        Hongpeng~Wang,
        Guangming~Lu,~\IEEEmembership{Senior Member, IEEE,}
        
        Jun~Yu,~\IEEEmembership{Senior Member, IEEE,}
        and Wenjie~Pei,~\IEEEmembership{Senior Member, IEEE}

        Harbin Institute of Technology, Shenzhen
        % <-this % stops a space
% \thanks{Jiagan~Huang,
%         Zikun~Zhou,
%         Hongpeng~Wang,
%         Guangming~Lu,
%         Jun~Yu,
%         and Wenjie~Pei are with Harbin Institute of Technology, Shenzhen, China (
%         e-mail: hjg1962781196@gmail.com; zhouzikunhit@gmail.com; wanghp@hit.edu.cn; 
% luguangm@hit.edu.cn; yujun@hit.edu.cn; wenjiecoder@outlook.com).

% Zikun~Zhou and Wenjie~Pei are the corresponding authors.}
}
% \thanks{This paper was produced by the IEEE Publication Technology Group. They are in Piscataway, NJ.}% <-this % stops a space
% \thanks{Manuscript received April 19, 2021; revised August 16, 2021.}}

% The paper headers
% \markboth{Journal of \LaTeX\ Class Files,~Vol.~14, No.~8, August~2021}%
% {Shell \MakeLowercase{\textit{et al.}}: A Sample Article Using IEEEtran.cls for IEEE Journals}

% \IEEEpubid{0000--0000/00\$00.00~\copyright~2021 IEEE}
% Remember, if you use this you must call \IEEEpubidadjcol in the second
% column for its text to clear the IEEEpubid mark.

\maketitle
\begin{abstract}

In Vision-Language Navigation (VLN) tasks involving multi-step instructions as guidance, an agent must associate its observations with the relevant instruction content at each decision point. Most graph-based methods commonly match candidate trajectories with the entire instruction, allowing already executed content to interfere with subsequent decisions. Although instruction-execution progress provides a direct cue for separating executed and pending content, existing progress estimation approaches often rely on explicit progress supervision, sophisticated progress-learning stages, or costly online reasoning with LLMs. In this paper, we propose a robust VLN framework with Awareness of Instruction-Execution Progress (VLN-AIEP), which couples progress estimation with navigation decision-making. VLN-AIEP associates the visited trajectory with the instruction to suppress embeddings corresponding to executed content, and then matches candidate locations with the pending content. This coupling makes correct navigation decisions depend on accurate progress estimation, allowing the progress estimator to learn directly from navigation decision labels without explicit progress annotations. To support progress estimation, we construct a landmark-anchored trajectory representation that encodes the semantic labels and viewpoint-relative geometry of instruction-referred landmarks. These landmarks serve as spatial references for associating the visited trajectory with executed instruction content. Extensive experiments on R2R, RxR, CVDN, and R2R-CE demonstrate that the proposed method achieves favorable performance across multiple metrics.
\end{abstract}

% !TeX root = ../bare_jrnl_new_sample4.tex
\section{Introduction}
\label{sec:intro}

Vision-Language Navigation (VLN)~\cite{anderson2018vision,li2024human,lin2025navcot,zhang2024navid,chen2024mapgpt} aims to empower an agent with visual sensors to navigate in a 3D environment following natural language instructions. For example, given an instruction like 
\textit{``Walk towards the French doors and turn left. Walk past the kitchen area...}'', the agent needs to comprehend this instruction, make movement decisions while perceiving its surroundings, and accurately halt at the destination. Recently, VLN has witnessed significant progress with the introduction of topological graphs~\cite{chen2022think,lu2024pret}. These graphs record all visited and navigable locations observed so far, thereby expanding the action space to the entire explored area and facilitating high-level planning over long horizons.
Despite these advances, following detailed multi-step instructions remains challenging, as the agent needs to accurately associate its observation with the relevant instruction to take an appropriate action at each critical decision point.

\begin{figure}
  \centering
   \includegraphics[width=1\linewidth]{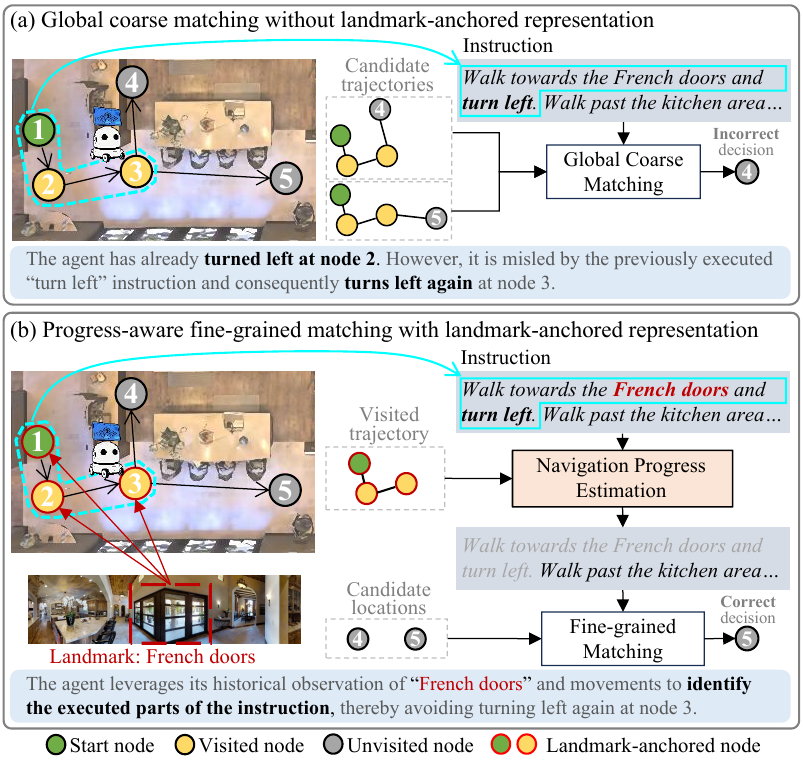}
   \caption{Illustration of our motivation. (a) Existing methods typically perform global coarse matching between candidate trajectories and the entire instruction. The agent is susceptible to interference from previously executed portions of the instruction, such as “turn left” in this case, leading to an incorrect decision. (b) Our method first estimates navigation progress by associating the visited trajectory with the instruction, where instruction-referred landmarks (\textit{e.g.}, ``French doors'') serve as spatial references with semantic and spatial cues. It then performs fine-grained matching with the pending instruction parts, thereby avoiding interference from executed instructions.
   }
   \label{fig:1}
   
\end{figure}

Representative graph-based methods~\cite{chen2022think,an2022bevbert,lu2024pret} typically construct candidate trajectories by appending candidate locations to the historical trajectory. Subsequently, they match the entire candidate trajectory with the entire instruction, \textit{i.e.},~global coarse matching, to determine the next movement. 
Such global coarse matching allows parts of the instruction that have already been executed to continue influencing the current decision. As illustrated in Figure~\ref{fig:1}(a), this interference can mislead the agent into executing ``\textit{turn left}'' twice. Intuitively, a multi-step instruction should be followed like a checklist: once a step is completed, subsequent decisions should be guided by the remaining instruction steps.
Accordingly, once the agent has turned left in Figure~\ref{fig:1}(b), it should shift its focus to locating the ``\textit{kitchen area}'' in subsequent navigation. Robust navigation therefore requires the agent to estimate instruction-execution progress, namely, how much of the instruction has been completed.

Several existing approaches~\cite{ma2019self,ma2019regretful,hong2020sub,he2021landmark,cui2026fca,guo2026aware,wang2026progressthink,long2024discuss,qiao2025open} have taken steps toward equipping agents with the ability to estimate instruction-execution progress. Traditional progress monitors~\cite{ma2019self,ma2019regretful} use normalized distance to the goal as supervision. However, this geometric progress signal can be inconsistent with actual instruction completion. To reduce this mismatch, some methods~\cite{hong2020sub,he2021landmark,cui2026fca,guo2026aware} introduce additional supervision to learn fine-grained instruction-trajectory alignment or progress awareness. However, whether manually annotated~\cite{hong2020sub,he2021landmark} or automatically generated~\cite{cui2026fca,guo2026aware}, such supervision requires additional annotation or a data-generation pipeline. Recently, Progress-Think~\cite{wang2026progressthink} circumvents the need for explicit progress annotations, but still requires dedicated progress-learning stages within a three-stage training procedure. Alternatively, LLM-based methods infer progress online from instructions and visual navigation context, but require repeated, costly inference~\cite{long2024discuss,qiao2025open}. Together, these trade-offs motivate us to learn progress estimation directly from navigation decision labels. Beyond supervision, accurate progress estimation requires associating trajectory states with instruction steps. Instruction-referred landmarks provide semantic and spatial cues that help establish this association. However, the use of such landmarks as spatial references for navigation progress estimation remains underexplored.

In this paper, we propose a robust VLN framework with Awareness of Instruction-Execution Progress (VLN-AIEP). To learn to estimate instruction-execution progress directly from navigation decision labels, VLN-AIEP is designed to couple progress estimation with navigation decision-making, making accurate progress estimation a prerequisite for correct navigation decisions. 
Specifically, it first associates the visited trajectory with the instruction to estimate execution progress and suppresses embeddings associated with executed content. It then matches candidate locations with the pending instruction content to make a navigation decision, as illustrated in Figure~\ref{fig:1}(b). This coupling allows navigation decision labels to supervise progress estimation without explicit progress annotations, while focusing on pending content mitigates interference from executed instruction content.
To further facilitate progress estimation, we construct a landmark-anchored trajectory representation that encodes the semantic labels and viewpoint-relative geometry of instruction-referred landmarks. These landmarks serve as spatial references for associating the visited trajectory with executed instruction content.

We evaluate our method on R2R~\cite{anderson2018vision}, RxR~\cite{ku2020room}, and CVDN~\cite{thomason2020vision}, and further validate its effectiveness in the continuous R2R-CE environment~\cite{krantz2020beyond}. VLN-AIEP performs competitively with state-of-the-art approaches in quantitative evaluations. In addition, qualitative analyses show that VLN-AIEP estimates navigation progress accurately. Our contributions can be summarized as:

\begin{itemize}
\item We propose VLN-AIEP, which couples progress estimation with navigation decision-making to focus on pending instruction content and learn to estimate progress from navigation labels without explicit progress supervision.

\item We construct a landmark-anchored trajectory representation that encodes semantic labels and viewpoint-relative geometry, allowing instruction-referred landmarks to serve as spatial references for progress estimation.

\item Extensive experiments on R2R, RxR, CVDN, and R2R-CE demonstrate favorable performance of VLN-AIEP against state-of-the-art approaches.
\end{itemize}

% !TeX root = ../bare_jrnl_new_sample4.tex
\section{Related Work}
\label{sec:formatting}
\vspace {1mm}
\noindent{\bf Vision-language navigation.}
VLN methods have evolved from sequence-based cross-modal learning to structured long-horizon reasoning. Early agents improved instruction-observation alignment through data augmentation and auxiliary objectives~\cite{fried2018speaker,wang2019reinforced,tan2019learning}. Transformer-based methods subsequently strengthened temporal modeling through multimodal pretraining and explicit history encoding~\cite{hao2020towards,qiao2022hop,chen2021history,lin2022multimodal,wang2023gridmm}. Building on these advances, graph-based methods explicitly represent the explored environment~\cite{chen2022think,an2022bevbert,lu2024pret,an2024etpnav}. These representations support local grounding and long-range planning. Continuous-environment methods further relax the assumption of a predefined navigation graph~\cite{krantz2020beyond,krantz2022sim,hong2022bridging}.
More recently, rapid advances in foundation models have led to their growing adoption in VLN. These methods generally follow two paradigms: online reasoning and task adaptation. Online agents invoke foundation models during navigation. They reason over textualized observations, maps, or multiple experts~\cite{zhou2024navgpt,chen2024mapgpt,long2024instructnav,long2024discuss,qiao2025open,xue2026profocus,dai2026history}. This enables flexible zero-shot decision-making but requires repeated inference. Adaptation-based systems instead train navigation policies or reasoning modules. They use video trajectories, generated supervision, or multi-task data~\cite{zhang2024navid,lin2025navcot,guo2026aware,Wei2025StreamVLNSV,zhang2026navfom,jiang2026m3e}. This reduces reliance on explicit online prompting but requires additional data and training.

Unlike representative graph-based methods that match complete candidate trajectories against the entire instruction, VLN-AIEP uses the visited trajectory to attenuate instruction embeddings associated with likely executed content. It then matches candidate-location embeddings with the resulting progress-aware instruction embeddings, which emphasize the pending instruction content. This progress-aware matching reduces the influence of previously executed instruction steps on the current decision.

\vspace{1mm}
\noindent{\bf Instruction-execution progress estimation.}
Existing methods obtain instruction-related progress information from geometric targets, fine-grained supervision, dedicated progress learning, or online reasoning. Early progress monitors use normalized goal distance to supervise progress estimation, and the Regretful Agent further uses the estimated progress for backtracking~\cite{ma2019self,ma2019regretful}. However, geometric progress may not reflect instruction completion because different instruction segments can cover unequal distances.
To better reflect instruction completion, fine-grained methods use human-annotated correspondences between instruction parts and navigation states. These annotations associate sub-instructions or landmarks with positions along the trajectory~\cite{hong2020sub,he2021landmark}, but are costly to collect. LAW similarly introduces language-aligned waypoint supervision~\cite{raychaudhuri2021law}, which requires a dedicated target-construction process. To reduce manual annotation, FCA-NIG and AwareVLN automatically construct alignment or progress-aware reasoning supervision before policy training~\cite{cui2026fca,guo2026aware}. 
Similarly, ATD uses a powerful LLM to generate progress-related state estimates for training its state-estimation branch~\cite{zhang2025cross}. Although more scalable, they still require a separate data-generation pipeline.
Progress-Think avoids explicit progress annotations through self-alignment~\cite{wang2026progressthink}. However, it employs a dedicated progress module, progress-specific objectives, and a three-stage training procedure.
Another line of work relies on online foundation-model reasoning to infer progress or subgoal completion. DiscussNav and Open-Nav reason over the instruction and navigation context, while SkillNav and Fast-SmartWay track the current subgoal~\cite{long2024discuss,qiao2025open,Ma2025BreakingDA,shi2025fast}. Although these methods are effective, they repeatedly invoke foundation models during navigation, resulting in considerable computational overhead and inference latency.

By contrast, VLN-AIEP couples progress estimation with navigation decision-making, making accurate progress estimation a prerequisite for correct decisions. This design allows navigation labels to supervise progress estimation without explicit progress annotations or repeated foundation-model inference.

\vspace{1mm}
\noindent{\bf Applications of landmarks in VLN.}
Objects often play a role similar to landmarks in VLN. Are You Looking? represents objects using labels, attributes, and bounding boxes, while ROAD incorporates object and room semantics~\cite{hu2019you,qi2021road}. DUET further includes object features in its fine-scale panoramic representation for local grounding~\cite{chen2022think}. These object cues enhance visual grounding but are not used as trajectory evidence for progress estimation.
Landmarks also provide fine-grained cross-modal supervision. Landmark-RxR introduces annotated correspondences between instruction segments and trajectory states, together with landmark-based rewards~\cite{he2021landmark}. GELA associates entity phrases with landmark regions through additional annotations and grounding objectives~\cite{cui2023gela}. In both methods, landmarks mainly serve as supervision targets. EXOR aligns landmark and direction relations across instructions and visual observations, but does not model landmark distances or their accumulation along the visited trajectory~\cite{zhang2022explicit}.
\begin{figure*}[t]
  \centering
   \includegraphics[width=1\linewidth]{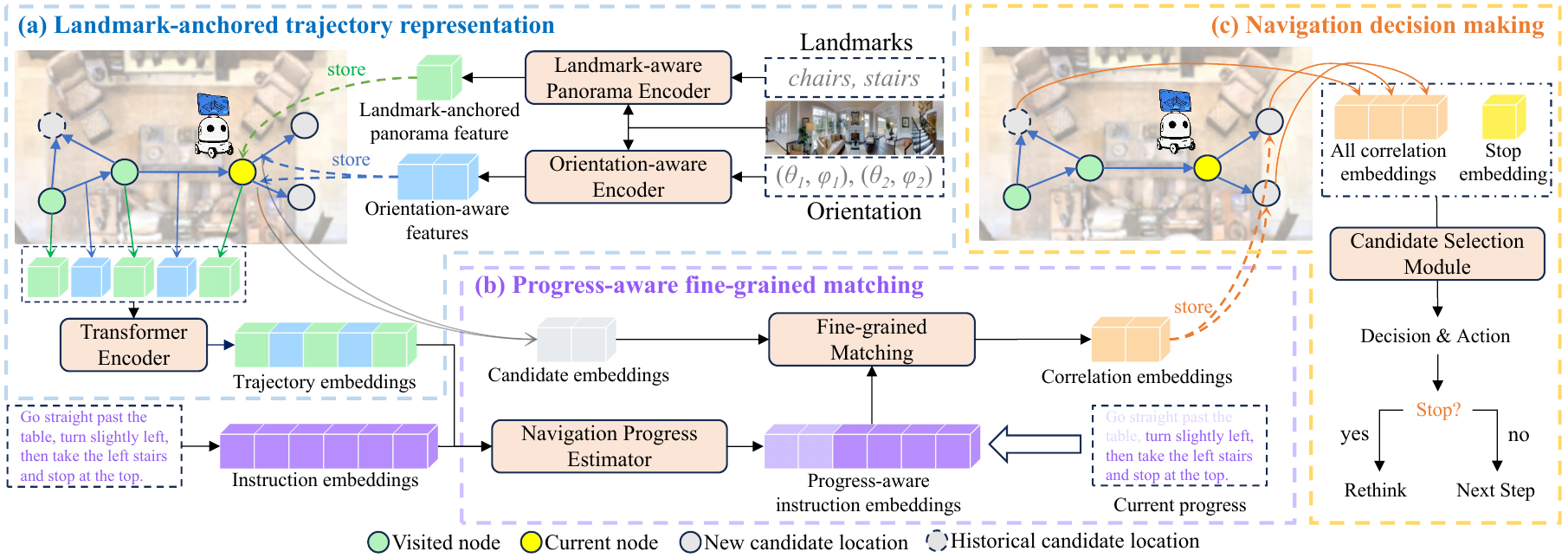}
   \caption {Overview of our method, which is based on a directed graph and comprises three components: (a) the landmark-anchored trajectory representation, (b) the progress-aware fine-grained matching module, and (c) the navigation decision-making module. Our method identifies the detour-free path from the initial node to the current node and then extracts the landmark-anchored trajectory representation of this path. With the historical trajectory embeddings, our method first estimates the navigation progress and then performs fine-grained matching to obtain the correlation embeddings for the new candidate locations. Finally, it makes a decision based on the correlation embeddings of both the new and historical candidate locations, as well as a stop embedding.}
   \label{fig:2}
\end{figure*}
Some methods further use landmarks or semantic constraints to monitor navigation stages. CONSOLE maintains a landmark pointer and forcibly advances it when the pointer remains unchanged for a predefined number of steps~\cite{lin2024correctable}. CA-Nav maintains completion constraints for the current instruction segment and uses minimum and maximum step thresholds to control stage transitions~\cite{chen2024CANav}. Such threshold-controlled monitoring is relatively rigid and may be less robust to variations in navigation execution and visual perception.
LLM-based methods such as DiscussNav and Open-Nav also extract landmarks from instructions and use them to support subsequent navigation decisions~\cite{long2024discuss,qiao2025open}.

Unlike methods that use landmarks as object cues, supervision targets, or discrete progress states, VLN-AIEP treats instruction-relevant landmarks as spatial anchors. It encodes their semantic identities and relative geometry into trajectory representations to support progress estimation.

% !TeX root = ../bare_jrnl_new_sample4.tex

\section{Method}

\subsection{Problem formulation}

For clarity, we formulate the VLN problem using the discrete setting~\cite{anderson2018vision,ku2020room,thomason2020vision} as a representative case. The environment is modeled as an undirected graph \( G = \{V, E\} \). Here, \( V = \{V_t\}_{t=1}^N \) is a set of \( N \) navigable nodes, and \( E \) denotes connectivity edges. Once receiving a natural language instruction, denoted by \(I\), the agent needs to understand the instruction and accordingly proceed to the destination. While positioned at node $V_t$, the agent observes three types of information: 1) a panorama \( P_t = \{p_{t,i}\}_{i=1}^K \) consisting of $K$ sub-view images, where \( p_{t,i} \) denotes the feature extracted from the $i^{th}$ sub-view image; 2) the set of navigable neighboring nodes \( N(V_t) \), where \( N(V_t) \subset V \); and 3) the orientations (consisting of heading and elevation) and coordinates of these navigable neighboring nodes. At each step, the agent decides to move to a navigable neighboring node or stop, based on the instruction and observation. Navigation is considered successful if the agent stops within 3 meters of the destination.

\subsection{Overall framework}

Figure~\ref{fig:2} presents the overall framework of VLN-AIEP, which couples progress estimation with navigation decision-making. This coupling enables progress awareness to be learned from navigation supervision without explicit progress annotations, while the estimated progress helps suppress interference from executed instruction content. VLN-AIEP comprises three main components: the landmark-anchored trajectory representation module, the progress-aware fine-grained matching module, and the navigation decision-making module. 

During the navigation process, VLN-AIEP utilizes a directed topological graph to record the visited nodes, the navigable nodes observed so far, and the directed connectivity between them. At each step, VLN-AIEP identifies the detour-free path from the initial node to the current node as the historical trajectory, and extracts the historical trajectory embeddings via the \textit{landmark-anchored trajectory representation module}. It treats the navigable nodes adjacent to its current location as new candidate locations. Each candidate location is represented by a candidate embedding extracted from the sub-view oriented toward it from the current node. 
With the trajectory, candidate, and instruction embeddings, VLN-AIEP leverages the \textit{progress-aware fine-grained matching module} to calculate the correlation between each candidate embedding and the pending instruction content. Specifically, the module suppresses executed instruction content based on the estimated progress and matches the resulting progress-aware instruction embeddings with candidate embeddings, producing the correlation embeddings for navigation decision-making.

Finally, the correlation embeddings of both the new and historical candidate locations, as well as a stop embedding, are fed into the \textit{navigation decision-making module} to make a decision. Here, incorporating the historical candidate locations into decision-making can effectively facilitate long-horizon backtracking. We next describe each component.

\subsection{Landmark-anchored trajectory representation}
\label{Landmark-anchored}

A historical trajectory representation is crucial for an agent to estimate its instruction-execution progress. Considering that navigation instructions often use landmarks as references to describe movements, we propose to enhance trajectory representations with landmarks, aiming to facilitate progress estimation. Specifically, we construct landmark-anchored trajectory representations by encoding the semantic labels and viewpoint-relative geometry of instruction-referred landmarks into panorama features, referred to as landmark-anchored panorama features. To further characterize movement direction, we interleave orientation-aware edge features with these panorama features in the trajectory representation, as shown in Figure~\ref{fig:2}(a). Next, we elaborate on this representation.

\vspace{1mm}
\noindent\textbf{Landmark-anchored panorama feature.} 
To identify the landmarks referenced in the instruction at node $V_t$, we first employ YOLOE~\cite{wang2025yoloe}, an open-vocabulary detection model, to process panoramic images using a vocabulary of common objects in the environment. With the detection results and depth maps, we extract a set of landmarks with various attributes, denoted by \( L_t = \left\{ ( l_i, \varphi_i, \theta_i, d_i, c_i ) \right\}_{i=1}^M \), where $M$ is the number of landmarks observed around $V_t$. For the $i^{th}$ landmark, \( l_i \) represents its textual name, $\varphi_i$ and $\theta_i$ are its heading and elevation \textit{w.r.t.} $V_t$, respectively, $d_i$ denotes the distance from the landmark to $V_t$, and $c_i$ is the detection confidence score. Based on detection confidence and instruction similarity, we select the $m$ most relevant landmarks to form $L_{\mathrm{ref}}$, which will be encoded into panorama features.

To enable each selected landmark $l_i \in L_{\mathrm{ref}}$ to serve as a spatial reference for progress estimation, we augment its textual embedding $f_i^l$ with relative orientation and distance cues, yielding the landmark feature $F_i^l$. Specifically, we first concatenate $f^l_i$ with the trigonometric representations of the orientation parameters \( \varphi_i \) and \( \theta_i \). Then, we map them to the visual space with a learnable projection matrix $W$, and fuse the projection result with the distance information $d_i$ of the landmark $l_i$. This process is formulated as:
\begin{equation}
  F^l_i = \left[ f^l_i ; \sin(\varphi_i), \cos(\varphi_i), \sin(\theta_i), \cos(\theta_i) \right] W \oplus \mathcal{P}(d_i),
\end{equation}
where \([ ; ]\) is the concatenation operation, $\oplus$ denotes element-wise addition, and \( \mathcal{P} \) is an MLP processing the distance $d_i$. Denote the feature sequence of the landmarks belonging to $L_{ref}$ by $F^L_{ref}$. We further integrate the landmark features $F^L_{ref}$ with the panorama features \( P_t \) via a Landmark-aware Panorama Encoder. This encoder consists of a cross-attention layer (in which $P_t$ is used as the query and $F^L_{ref}$ is used as the keys and values) with residual fusion and two Transformer encoding layers. The resulting landmark-anchored panorama features \(N_t\) of node $V_t$ can be formulated as: 
\begin{equation}
  N_t = \phi_{enc} \left(P_t \oplus \phi_{ca}(P_t, F^{L}_{ref})\right),
\end{equation}
where $\phi_{ca}(\cdot, \cdot)$ denotes the above-mentioned cross-attention layer, and $\phi_{enc}$ denotes the two Transformer encoding layers.

\vspace{1mm}
\noindent
\textbf{Orientation-aware feature.} Following~\cite{lu2024pret}, we extract an orientation-aware feature for each directed connection from $V_t$ to an adjacent node. The encoder uses the relative-orientation embeddings of these connections as queries and the panorama features as keys and values in a cross-attention layer. The resulting features are stored on the corresponding edges.

\vspace{1mm}
\noindent
\textbf{Trajectory representation.} With the landmark-anchored panorama features stored on visited nodes and the orientation-aware features stored on directed edges, we construct a representation of the historical trajectory. Specifically, we take the detour-free path from the initial node to the current node as the historical trajectory, and interleave the node and edge features in traversal order. The resulting feature sequence is fed into a Transformer encoder to produce the trajectory embeddings $T_e$, which capture both instruction-referred landmarks and movement directions.

\subsection{Progress-aware fine-grained matching} 
\label{pam}

To enable progress-aware navigation and facilitate learning to estimate instruction-execution progress, we propose a progress-aware fine-grained matching module. As shown in Figure~\ref{fig:2}(b), this module comprises a navigation progress estimator and a fine-grained matching block. 
Given the historical trajectory embeddings $T_e$ and instruction embeddings $I_e$, the navigation progress estimator suppresses instruction content that is likely to have been executed, producing progress-aware instruction embeddings that emphasize the pending content. The new candidate embeddings are then matched only with these progress-aware instruction embeddings, thereby reducing interference from executed content. This design makes candidate evaluation explicitly depend on the estimated instruction-execution progress, allowing navigation supervision to provide a learning signal for progress estimation without explicit progress labels. We detail each component below.

\begin{figure}[t]
  \centering
   \includegraphics[width=0.85\linewidth]{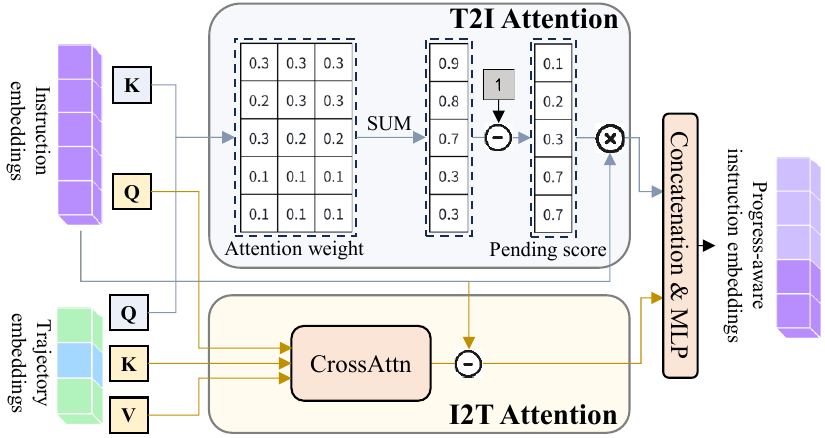}
   \caption{Structure of the navigation progress estimator. It employs the instruction-to-trajectory (I2T) and trajectory-to-instruction (T2I) attentions to suppress the instruction embeddings that resemble the trajectory embeddings, yielding progress-aware instruction embeddings that indicate instruction-execution progress.}
   \label{fig:3}
\end{figure}

\vspace{1mm}
\noindent
\textbf{Navigation progress estimator.}
Given that explicitly segmenting an instruction into executed and pending parts is inherently challenging, we estimate instruction-execution progress through embedding modulation. 
Specifically, the estimator uses the similarity between the instruction and visited trajectory embeddings to identify likely executed content and suppresses the corresponding instruction embeddings.
As shown in Figure~\ref{fig:3}, the estimator consists of Trajectory-to-Instruction (T2I) attention and Instruction-to-Trajectory (I2T) attention. The two branches suppress executed instruction content in different ways, and their outputs are fused to produce the progress-aware instruction embeddings.

In the T2I attention branch, we first calculate attention weights $A_{\text{T2I}}$ from the trajectory embeddings \(T_e\) to the instruction embeddings \(I_e\). These weights are then summed along the dimension corresponding to \(T_e\) to obtain \(S_I\), which represents the executed degree of each instruction embedding. Values of $S_I$ greater than 1 are clipped to 1. We subsequently employ $1-S_I$, referred to as the \textit{pending score}, to modulate the instruction embeddings, thereby attenuating those more likely to correspond to already executed segments. The resulting instruction embeddings $I_e^{\text{T2I}}$ can be formulated as:
\begin{equation}
\begin{split}
&S_I[i] = \sum_{j=1}^{len(T_e)} A_\text{T2I}[i, j], \\
&I_e^{\text{T2I}} = (1 - S_I) \odot I_e.
\end{split}
\end{equation}

The I2T attention branch employs a cross-attention layer to extract the components of \( T_e \) that are highly relevant to \( I_e \), where \( I_e \) is used as the query. The cross-attention result is then subtracted from the original \( I_e \) to obtain \( I_e^{\text{I2T}} \), which primarily suppresses the parts of \( I_e \) that match \( T_e \). The above operations can be formulated as:

\begin{equation}
I_e^{\text{I2T}} = I_e \ominus \phi_{ca}^{\text{I2T}}(I_e, T_e),
\end{equation}
where $\phi_{ca}^{\text{I2T}}$ denotes the above-mentioned cross-attention layer, and $\ominus$ denotes element-wise subtraction.
To fuse \(I_e^{\text{I2T}}\) and \(I_e^{\text{T2I}}\), we first concatenate these two embeddings along the feature dimension. Subsequently, we project the concatenated result back to the original feature dimension through an MLP, producing the progress-aware instruction embeddings \(I_e^{pa}\).

\vspace{1mm}
\noindent
{\bf Fine-grained matching.} With the progress-aware instruction embeddings $I_e^{pa}$, we introduce a matching block to generate correlation embeddings that characterize the consistency between each candidate embedding and the pending instruction content. Technically, the matching block is implemented as a Transformer decoder where candidate embeddings serve as queries. 
This design restricts the attention scope to the new candidates and the modulated instructions, avoiding repeated processing of the historical trajectory during matching.

\vspace{1mm}
\noindent{\bf Implicit progress supervision through navigation loss.}
The navigation progress estimator is trained without explicit progress labels. During fine-grained matching, the correlation embeddings of new candidates are computed only from \(I_e^{pa}\) and their candidate embeddings, without reintroducing \(T_e\). Correct candidate matching therefore depends on whether \(I_e^{pa}\) suppresses executed content while preserving pending content. An inaccurate progress estimate may suppress content that remains pending or retain content that has already been executed, thereby weakening the correlation with the correct candidate and increasing the navigation loss. During end-to-end training, gradients from this loss propagate through the matching block to the progress estimator. Consequently, conventional navigation decision labels provide an implicit learning signal for instruction-execution progress estimation.

\subsection{Navigation decision-making}
The correlation embeddings generated by the fine-grained matching block are fed into a candidate selection module for navigation decision-making, as shown in Figure~\ref{fig:2}(c). The module comprises an MLP layer, a Transformer encoder layer, and a softmax output layer, similar to~\cite{lu2024pret}. To enable the agent to make a termination decision, we append a learnable stop embedding to the correlation embeddings. The module then outputs a probability distribution over all candidate locations and the stop action. To improve the robustness of decision-making, we further introduce the following mechanisms.

\vspace{1mm}
\noindent
{\bf Incorporating historical candidates into decision-making.}
To facilitate long-horizon backtracking, we incorporate historical candidates alongside the new candidates adjacent to the current node, as illustrated in Figure~\ref{fig:2}(c). If the agent continues navigation, the correlation embeddings of unselected new candidates are stored on the corresponding graph nodes. At subsequent steps, these locations serve as historical candidates, with their stored embeddings preserving their consistency with the pending instruction content when first observed. If a historical candidate is selected, the agent navigates to it along the shortest path in the constructed graph.

\vspace{1mm}
\noindent
{\bf Rethinking the termination decision.}
The above mechanism expands the action space over unvisited locations, whereas visited locations remain masked during decision-making. Consequently, the agent cannot return to a previously visited location if it has moved past the appropriate stopping point. We therefore introduce a rethinking strategy that re-ranks the last \(n\) visited locations using the candidate selection module once the stop action is selected. The highest-scoring location is selected as the final stopping location, allowing the agent to return to a more appropriate location before terminating.

\subsection{Model learning}

To effectively integrate landmarks with the trajectory representation, we propose landmark-anchored Masked Language Modeling (MLM) for model optimization. Unlike standard MLM~\cite{devlin2019bert} that masks tokens randomly, our method specifically masks text tokens corresponding to landmarks. 
This objective encourages the model to use visual trajectory features when predicting landmark tokens, thereby strengthening the association between landmark tokens and visual trajectory features.
In practice, we follow the training paradigm of PRET~\cite{lu2024pret}, consisting of pretraining and fine-tuning stages. In the pretraining stage, we use our landmark-anchored MLM alongside the standard MLM. Subsequently, we combine teacher-forcing and student-forcing to fine-tune the model.

\begin{table*}[t!]
\centering
\setlength{\tabcolsep}{6.6pt}
\renewcommand{\arraystretch}{1.1}
\caption{Experimental results of VLN methods on R2R~\cite{anderson2018vision} dataset. $\downarrow$ and $\uparrow$ indicate that lower/higher values are preferred, respectively. Methods marked with $^{\blacklozenge}$ are based on large models. ``Progress awareness'' indicates whether navigation progress is estimated.}
\vspace{-2mm}
\resizebox{0.975\linewidth}{!}{
  \begin{tabular}{llcccccccccccccccc} 
  \toprule
  \multirow{2}{*}{Methods} & \multirow{2}{*}{Publication} &
  \multirow{2}{*}{\makecell{Progress\\awareness}} 
  && \multicolumn{4}{c}{Val Seen} && \multicolumn{4}{c}{Val Unseen} && \multicolumn{4}{c}{Test Unseen} \\ 
  \cline{5-8} \cline{10-13} \cline{15-18}
  &&&& TL & NE$\downarrow$ & SR$\uparrow$ & SPL$\uparrow$ && TL & NE$\downarrow$ & SR$\uparrow$ & SPL$\uparrow$ && TL & NE$\downarrow$ & SR$\uparrow$ & SPL$\uparrow$ \\ 
  \midrule
  Seq2Seq-SF~\cite{anderson2018vision} & CVPR'18 & No && 11.33 & 6.01 & 39 & - && 8.39 & 7.81 & 22 & - && 8.13 & 7.85 & 28 & 18 \\
  Self-Monitoring~\cite{ma2019self} & ICLR'19 & Yes && - & 3.22 & 67 & 58 && - & 5.52 & 45 & 32 && - & 5.99 & 43 & 32 \\
  Regretful~\cite{ma2019regretful} & CVPR'19 & Yes && - & 3.23 & 69 & 63 && - & 5.32 & 50 & 41 && - & 5.69 & 56 & 40 \\
  HAMT~\cite{chen2021history} & NIPS'21 & No && 11.15 & 2.51 & 76 & 72 && 11.46 & \textbf{2.29} & 66 & 61 && 12.27 & 3.93 & 65 & 60 \\
  DUET~\cite{chen2022think} & CVPR'22 & No && 12.32 & 2.28 & 79 & 73 && 13.94 & 3.31 & 72 & 60 && 14.73 & 3.65 & 69 & 59 \\
  BEVBert~\cite{an2022bevbert} & ICCV'23 & No && 13.56 & 2.17 & \textbf{81} & 74 && 14.55 & 2.81 & 75 & 64 && 15.87 & 3.13 & 73 & 62 \\
  PRET~\cite{lu2024pret} & ECCV'24 & No && 11.25 & 2.41 & 78 & 72 && 11.87 & 2.90 & 74 & 65 && 12.21 & 3.09 & 72 & \textbf{64} \\
\cellcolor{white}DiscussNav$^{\blacklozenge}$~\cite{long2024discuss}
& \cellcolor{white}ICRA'24
& \cellcolor{white}Yes
& \cellcolor{white}
& \cellcolor{white}-
& \cellcolor{white}-
& \cellcolor{white}-
& \cellcolor{white}-
& \cellcolor{white}
& \cellcolor{white}9.69
& \cellcolor{white}5.32
& \cellcolor{white}43
& \cellcolor{white}40
& \cellcolor{white}
& \cellcolor{white}-
& \cellcolor{white}-
& \cellcolor{white}-
& \cellcolor{white}- \\

\cellcolor{white}ATD$^{\blacklozenge}$~\cite{zhang2025cross}
& \cellcolor{white}CVPR'26
& \cellcolor{white}Yes
& \cellcolor{white}
& \cellcolor{white}12.26
& \cellcolor{white}2.67
& \cellcolor{white}76
& \cellcolor{white}67
& \cellcolor{white}
& \cellcolor{white}13.33
& \cellcolor{white}2.81
& \cellcolor{white}75
& \cellcolor{white}63
& \cellcolor{white}
& \cellcolor{white}14.00
& \cellcolor{white}3.03
& \cellcolor{white}\textbf{74}
& \cellcolor{white}63 \\
  
  \multicolumn{1}{>{\cellcolor{mycolor}}l}{\textbf{VLN-AIEP(Ours)}}  
    & \multicolumn{1}{>{\cellcolor{mycolor}}c}{}
  & \multicolumn{1}{>{\cellcolor{mycolor}}c}{Yes}
  & \multicolumn{1}{>{\cellcolor{mycolor}}c}{}
  & \multicolumn{1}{>{\cellcolor{mycolor}}c}{11.71}  
  & \multicolumn{1}{>{\cellcolor{mycolor}}c}{\textbf{2.17}}  
  & \multicolumn{1}{>{\cellcolor{mycolor}}c}{\textbf{81}}  
  & \multicolumn{1}{>{\cellcolor{mycolor}}c}{\textbf{75}}  
  & \multicolumn{1}{>{\cellcolor{mycolor}}c}{}  
  & \multicolumn{1}{>{\cellcolor{mycolor}}c}{13.28}  
  & \multicolumn{1}{>{\cellcolor{mycolor}}c}{2.73}  
  & \multicolumn{1}{>{\cellcolor{mycolor}}c}{\textbf{77}}  
  & \multicolumn{1}{>{\cellcolor{mycolor}}c}{\textbf{66}}  
  & \multicolumn{1}{>{\cellcolor{mycolor}}c}{}  
  & \multicolumn{1}{>{\cellcolor{mycolor}}c}{13.70} 
  & \multicolumn{1}{>{\cellcolor{mycolor}}c}{\textbf{2.91}} 
  & \multicolumn{1}{>{\cellcolor{mycolor}}c}{\textbf{74}} 
  & \multicolumn{1}{>{\cellcolor{mycolor}}c}{\textbf{64}} \\  
  \bottomrule
  \end{tabular}
}
\label{tab:r2r_sota}
\vspace{-4mm}
\end{table*}

\begin{table}[t]
\centering
\setlength{\tabcolsep}{5pt}
\renewcommand{\arraystretch}{1.1}
\caption{Experimental results of the VLN methods on RxR~\cite{ku2020room} dataset. ``Progress awareness'' indicates whether navigation progress is estimated.} 
\vspace{-2mm}
\resizebox{.98\linewidth}{!}{
\begin{tabular}{lccccccccc}
\toprule
\multirow{2}{*}{Methods}
& \multirow{2}{*}{\makecell{Progress\\awareness}}
&& \multicolumn{3}{c}{Val Seen}
&& \multicolumn{3}{c}{Val Unseen} \\
\cline{4-6} \cline{8-10}
&&& NE$\downarrow$ & SR$\uparrow$ & sDTW$\uparrow$
&& NE$\downarrow$ & SR$\uparrow$ & sDTW$\uparrow$ \\
\midrule
LSTM~\cite{ku2020room}          & No  && 10.7 & 25.2 & 20.7 && 10.9 & 22.8 & 18.2 \\
HAMT~\cite{chen2021history}     & No  && -    & 59.4 & 50.9 && -    & 56.5 & 48.3 \\
MARVEL~\cite{wang2022less}      & No  && 3.0  & 75.9 & 68.8 && 4.5  & 64.8 & 57.5 \\
BEVBert~\cite{an2022bevbert}    & No  && 3.2  & 75.0 & 66.7 && 4.0  & 68.5 & 58.6 \\
PRET~\cite{lu2024pret}          & No  && 2.4  & 79.3 & 70.7 && 3.2  & 72.8 & 62.4 \\
\multicolumn{1}{>{\cellcolor{mycolor}}l}{\textbf{VLN-AIEP(Ours)}}
& \multicolumn{1}{>{\cellcolor{mycolor}}c}{Yes}
& \multicolumn{1}{>{\cellcolor{mycolor}}c}{}
& \multicolumn{1}{>{\cellcolor{mycolor}}c}{\textbf{1.7}}
& \multicolumn{1}{>{\cellcolor{mycolor}}c}{\textbf{85.9}}
& \multicolumn{1}{>{\cellcolor{mycolor}}c}{\textbf{77.9}}
& \multicolumn{1}{>{\cellcolor{mycolor}}c}{}
& \multicolumn{1}{>{\cellcolor{mycolor}}c}{\textbf{2.9}}
& \multicolumn{1}{>{\cellcolor{mycolor}}c}{\textbf{74.5}}
& \multicolumn{1}{>{\cellcolor{mycolor}}c}{\textbf{62.9}} \\
\bottomrule
\end{tabular}}
\label{tab:rxr_sota}
\vspace{-2mm}
\end{table}

\begin{table}[t]
\centering
\setlength{\tabcolsep}{5.5pt}
\renewcommand{\arraystretch}{1.1}
\caption{Goal Progress (GP, $\uparrow$) of the VLN methods on CVDN~\cite{thomason2020vision} dataset. Methods marked with $^{\blacklozenge}$ are based on large models. ``Progress awareness'' indicates whether navigation progress is estimated.}
\vspace{-2mm}
\resizebox{.95\linewidth}{!}{
\begin{tabular}{lccccc}
\toprule
Methods & \makecell{Progress\\awareness} & Val Seen & Val Unseen & Test Unseen & Params \\
\midrule
HAMT~\cite{chen2021history}
& No & 6.91 & 5.13 & 5.58 & - \\

VLN-PETL~\cite{qiao2023vln}
& No & 7.76 & 5.69 & 6.13 & - \\

\cellcolor{white}NaviLLM$^{\blacklozenge}$~\cite{zheng2024towards}
& \cellcolor{white}No
& \cellcolor{white}-
& \cellcolor{white}6.16
& \cellcolor{white}\textbf{7.90}
& \cellcolor{white}$\sim 7.4B$ \\

\multicolumn{1}{>{\cellcolor{mycolor}}l}{\textbf{VLN-AIEP(Ours)}}
& \multicolumn{1}{>{\cellcolor{mycolor}}c}{Yes}
& \multicolumn{1}{>{\cellcolor{mycolor}}c}{\textbf{8.06}}
& \multicolumn{1}{>{\cellcolor{mycolor}}c}{\textbf{6.44}}
& \multicolumn{1}{>{\cellcolor{mycolor}}c}{6.90}
& \multicolumn{1}{>{\cellcolor{mycolor}}c}{$\mathbf{<} 0.5B$} \\
\bottomrule
\end{tabular}}
\label{tab:cvdn}
\vspace{-3mm}
\end{table}

\begin{table}[t]
\centering
\setlength{\tabcolsep}{4.5pt}
\renewcommand{\arraystretch}{1.1}
\caption{Experimental results on val-unseen split of R2R-CE~\cite{krantz2020beyond} dataset. Methods marked with $^{\blacklozenge}$ are based on large models. ``Progress awareness'' indicates whether navigation progress is estimated.}
\vspace{-2mm}
\resizebox{.98\linewidth}{!}{
\begin{tabular}{l c c cccc}
\toprule
Methods & Publication & \makecell{Progress\\awareness} & NE$\downarrow$ & OSR$\uparrow$ & SR$\uparrow$ & SPL$\uparrow$ \\
\midrule
Seq2Seq~\cite{krantz2020beyond} & ECCV'20 & No & 7.37 & 40 & 32 & 30 \\
BEVBert~\cite{an2022bevbert} & ICCV'23 & No & 4.57 & 67 & 59 & 50 \\
ETPNav~\cite{an2024etpnav} & TPAMI'24 & No & 4.71 & 65 & 57 & 49 \\
\cellcolor{white}Open-Nav$^{\blacklozenge}$~\cite{qiao2025open}
& \cellcolor{white}ICRA'25
& \cellcolor{white}Yes
& \cellcolor{white}6.70
& \cellcolor{white}23
& \cellcolor{white}19
& \cellcolor{white}16 \\

\cellcolor{white}Fast-SmartWay$^{\blacklozenge}$~\cite{shi2025fast}
& \cellcolor{white}ICRA'26
& \cellcolor{white}Yes
& \cellcolor{white}7.72
& \cellcolor{white}-
& \cellcolor{white}28
& \cellcolor{white}25 \\

\cellcolor{white}StreamVLN$^{\blacklozenge}$~\cite{Wei2025StreamVLNSV}
& \cellcolor{white}ICRA'26
& \cellcolor{white}No
& \cellcolor{white}4.98
& \cellcolor{white}64
& \cellcolor{white}57
& \cellcolor{white}52 \\

\cellcolor{white}DecoVLN$^{\blacklozenge}$~\cite{xin2026deco}
& \cellcolor{white}CVPR'26
& \cellcolor{white}No
& \cellcolor{white}5.01
& \cellcolor{white}64
& \cellcolor{white}56
& \cellcolor{white}51 \\

\cellcolor{white}Progress-Think$^{\blacklozenge}$~\cite{wang2026progressthink}
& \cellcolor{white}CVPR'26
& \cellcolor{white}Yes
& \cellcolor{white}4.68
& \cellcolor{white}64
& \cellcolor{white}\textbf{60}
& \cellcolor{white}\textbf{54} \\
\multicolumn{1}{>{\cellcolor{mycolor}}l}{\textbf{VLN-AIEP(Ours)}} &
\multicolumn{1}{>{\cellcolor{mycolor}}c}{} &
\multicolumn{1}{>{\cellcolor{mycolor}}c}{Yes} &
\multicolumn{1}{>{\cellcolor{mycolor}}c}{\textbf{4.39}} &
\multicolumn{1}{>{\cellcolor{mycolor}}c}{\textbf{68}} &
\multicolumn{1}{>{\cellcolor{mycolor}}c}{\textbf{60}} &
\multicolumn{1}{>{\cellcolor{mycolor}}c}{50} \\
\bottomrule
\end{tabular}}
\label{tab:r2r_ce_unseen}
\vspace{-3mm}
\end{table}

\begin{table*}[t]
\centering
\setlength{\tabcolsep}{11pt}  
\renewcommand{\arraystretch}{1.1}  
\caption{Ablation study on the core designs of VLN-AIEP. \textbf{PAM} denotes progress-aware fine-grained matching; \textbf{OA} denotes orientation-aware. }
\vspace{-2mm}
\resizebox{0.7\linewidth}{!}{
\begin{tabular}{llcccc}
\toprule
Trajectory representation & PAM & TL & NE$\downarrow$ & SR$\uparrow$ & SPL$\uparrow$ \\
\midrule
\multicolumn{1}{>{\cellcolor{white}}l}{1) Panorama features}   & \multicolumn{1}{>{\cellcolor{white}}l}{\textbf{ ~×}} & \multicolumn{1}{>{\cellcolor{white}}l}{15.83} & \multicolumn{1}{>{\cellcolor{white}}l}{3.74} & \multicolumn{1}{>{\cellcolor{white}}l}{67.39} & \multicolumn{1}{>{\cellcolor{white}}l}{56.45} \\
\multicolumn{1}{>{\cellcolor{grey1}}l}{2) Panorama \& OA features}     & \multicolumn{1}{>{\cellcolor{grey1}}l}{\textbf{ ~×}} & \multicolumn{1}{>{\cellcolor{grey1}}l}{13.68} & \multicolumn{1}{>{\cellcolor{grey1}}l}{3.12} & \multicolumn{1}{>{\cellcolor{grey1}}l}{72.92} & \multicolumn{1}{>{\cellcolor{grey1}}l}{62.44}\\
\multicolumn{1}{>{\cellcolor{mycolor}}l}{3) Landmark-anchored panorama \& OA features}  & \multicolumn{1}{>{\cellcolor{mycolor}}l}{\textbf{ ~×}} & \multicolumn{1}{>{\cellcolor{mycolor}}l}{13.31} & \multicolumn{1}{>{\cellcolor{mycolor}}l}{2.88} & \multicolumn{1}{>{\cellcolor{mycolor}}l}{74.15} & \multicolumn{1}{>{\cellcolor{mycolor}}l}{64.05} \\

\multicolumn{1}{>{\cellcolor{white}}l}{4) Panorama features}   & \multicolumn{1}{>{\cellcolor{white}}l}{~~\checkmark} & \multicolumn{1}{>{\cellcolor{white}}l}{13.91} & \multicolumn{1}{>{\cellcolor{white}}l}{3.32} & \multicolumn{1}{>{\cellcolor{white}}l}{69.47} & \multicolumn{1}{>{\cellcolor{white}}l}{60.02} \\

\multicolumn{1}{>{\cellcolor{grey1}}l}{5) Panorama \& OA features}     & \multicolumn{1}{>{\cellcolor{grey1}}l}{~~\checkmark} & \multicolumn{1}{>{\cellcolor{grey1}}l}{13.44} & \multicolumn{1}{>{\cellcolor{grey1}}l}{2.85} & \multicolumn{1}{>{\cellcolor{grey1}}l}{74.02} & \multicolumn{1}{>{\cellcolor{grey1}}l}{63.82} \\

\multicolumn{1}{>{\cellcolor{mycolor}}l}{6) Landmark-anchored panorama \& OA features} & 
\multicolumn{1}{>{\cellcolor{mycolor}}l}{~~\checkmark} &
\multicolumn{1}{>{\cellcolor{mycolor}}l}{13.21} & \multicolumn{1}{>{\cellcolor{mycolor}}l}{2.76} & \multicolumn{1}{>{\cellcolor{mycolor}}l}{76.40} & 
\multicolumn{1}{>{\cellcolor{mycolor}}l}{65.73} \\
\bottomrule
\end{tabular}}
\vspace{-4mm}
\label{tab:overall}
\end{table*}

\begin{table}[t]
\centering
\setlength{\tabcolsep}{10pt}
\renewcommand{\arraystretch}{1.1}
\caption{Ablation study on combining I2T and T2I attention.
}
\vspace{-2mm}
\resizebox{.95\linewidth}{!}{
\begin{tabular}{lcccc}
\toprule
Variants & TL & NE$\downarrow$ & SR$\uparrow$ & SPL$\uparrow$ \\
\midrule
1) \(w/o\) I2T attention  & 12.71 & 3.04 & 72.62 & 63.30 \\
2) \(w/o\) T2I attention  & 12.41 & 2.85 & 73.81 & 64.92  \\
\multicolumn{1}{>{\cellcolor{mycolor}}l}{3) full model (I2T+T2I)}   & \multicolumn{1}{>{\cellcolor{mycolor}}l}{13.21} & \multicolumn{1}{>{\cellcolor{mycolor}}l}{2.76} & \multicolumn{1}{>{\cellcolor{mycolor}}l}{76.40} &
\multicolumn{1}{>{\cellcolor{mycolor}}l}{65.73} \\
\bottomrule
\end{tabular}}
\label{tab:attention}
\vspace{-2mm}
\end{table}

\section{Experiments}
\subsection{Datasets and evaluation metrics}

\noindent{\bf Datasets. }The experiments are conducted on four datasets. 
R2R~\cite{anderson2018vision} is the standard benchmark for evaluating instruction-guided navigation. RxR~\cite{ku2020room} features longer paths and more detailed instructions, testing whether VLN-AIEP can follow long-horizon guidance. CVDN~\cite{thomason2020vision} provides short and incomplete dialog guidance, testing progress-aware decision-making under partial instructions. Finally, R2R-CE~\cite{krantz2020beyond} transfers R2R to continuous environments with noisy waypoint prediction, testing robustness beyond ideal graph transitions.

\vspace {1mm}
\noindent{\bf Evaluation metrics. }As in~\cite{anderson2018vision,ku2020room,krantz2020beyond}, we employ the following evaluation metrics: Trajectory Length (TL), Navigation Error (NE), Success Rate (SR): proportion of navigations with NE less than 3 meters, Oracle SR (OSR), Success Rate penalized by Path Length (SPL), and Success weighted by normalized Dynamic Time Warping (sDTW). For CVDN~\cite{thomason2020vision}, we adopt Goal Progress (GP), which quantifies the average progress toward the target.

\subsection{Implementation details}
\noindent{\bf Module architecture and hyperparameters. }As in~\cite{anderson2018vision,ku2020room,krantz2020beyond}, we use a 6-layer Transformer encoder initialized with the pretrained weights of ALBEF~\cite{li2021align} as the text backbone. For the multilingual dataset~\cite{ku2020room}, we instead use a 12-layer mRoBERTa~\cite{conneau2019unsupervised} as the text backbone. Following~\cite{lu2024pret}, we extract the image features using DINOv2~\cite{oquab2023dinov2}. We set  hyperparameters \( m \) and \( n \) to 20 and 2, respectively.

\vspace{1mm}

\noindent{\bf Training details.} For R2R, we pretrain VLN-AIEP with batch size 32, then fine-tune it with batch size 16. The augmented dataset~\cite{hao2020towards} is used for both stages, following prior work~\cite{chen2022think,an2022bevbert,lu2024pret}. For RxR, given its longer instructions and trajectories, we use a batch size of 8 for both pretraining and fine-tuning, and adopt Marky~\cite{wang2022less} as augmented data. For CVDN, we pretrain VLN-AIEP with batch size 64 and fine-tune with 16. For R2R-CE, we initialize VLN-AIEP using the weights trained on R2R, followed by fine-tuning on R2R-CE with a batch size of 8. 
All experiments use the AdamW optimizer~\cite{loshchilov2017decoupled}.

\subsection{Comparison with state-of-the-art methods}
Since DiscussNav~\cite{long2024discuss}, Open-Nav~\cite{qiao2025open}, and Fast-SmartWay~\cite{shi2025fast} follow a zero-shot setting and report results only on selected dataset splits, we report their results for reference only and do not directly compare them with ours.

\vspace {1mm}
\noindent{\bf Performance on R2R. }
Table~\ref{tab:r2r_sota} compares VLN-AIEP with previous methods on R2R. VLN-AIEP achieves SR scores of 81\%, 77\%, and 74\% on val-seen, val-unseen, and test-unseen, exceeding PRET~\cite{lu2024pret} by 3, 3, and 2 percentage points, respectively. Compared with the global coarse matching used by PRET, our progress-aware fine-grained matching reduces interference from executed instruction content, enabling more robust navigation across seen and unseen environments.

\vspace {1mm}
\noindent{\bf Performance on RxR. }
Table~\ref{tab:rxr_sota} reports the results on RxR. VLN-AIEP achieves SR scores of 85.9\% and 74.5\% on val-seen and val-unseen, exceeding PRET~\cite{lu2024pret} by 6.6 and 1.7 percentage points, respectively. These results indicate that our progress-aware fine-grained matching remains effective on longer paths with more complex instructions.

\vspace {1mm}
\noindent{\bf Performance on CVDN. }
Table~\ref{tab:cvdn} reports the results on CVDN. VLN-AIEP achieves GP scores of 8.06, 6.44, and 6.90 on val-seen, val-unseen, and test-unseen, respectively. With fewer than $0.5B$ parameters, it reaches 6.44 on val-unseen compared with 6.16 for NaviLLM~\cite{zheng2024towards} ($\sim 7.4B$), although its test-unseen score remains below NaviLLM's 7.90. These results demonstrate competitive performance under short and incomplete dialog guidance.

\vspace {1mm}
\noindent{\bf Performance on R2R-CE. }
Table~\ref{tab:r2r_ce_unseen} reports the val-unseen results on R2R-CE. VLN-AIEP achieves an NE of 4.39, an OSR of 68\%, an SR of 60\%, and an SPL of 50\%. With the same waypoint predictor~\cite{hong2022bridging}, it exceeds ETPNav~\cite{an2024etpnav} by 3 percentage points in both SR and OSR and by 1 percentage point in SPL. Although continuous environments more closely resemble real-world scenarios and introduce greater navigation noise, VLN-AIEP can still accurately estimate instruction execution progress to guide navigation.

\begin{table}[t]
\centering
\setlength{\tabcolsep}{8pt}
\renewcommand{\arraystretch}{1.1}
\caption{Ablation study on landmark-anchored MLM.}
\vspace{-2mm}
\resizebox{.95\linewidth}{!}{
\begin{tabular}{lcccc}
\toprule
Variants & TL & NE$\downarrow$ & SR$\uparrow$ & SPL$\uparrow$ \\
\midrule
1) \(w/o\) landmark-anchored MLM         & 12.64 & 3.02 & 72.32 & 63.77 \\
\multicolumn{1}{>{\cellcolor{mycolor}}l}{2) \(w/\) landmark-anchored MLM}     
& \multicolumn{1}{>{\cellcolor{mycolor}}l}{13.21} & \multicolumn{1}{>{\cellcolor{mycolor}}l}{2.76} & \multicolumn{1}{>{\cellcolor{mycolor}}l}{76.40} &
\multicolumn{1}{>{\cellcolor{mycolor}}l}{65.73} \\
\bottomrule
\end{tabular}}
\label{tab:MLM}
\end{table}

\begin{table}[t]
\centering
\setlength{\tabcolsep}{15pt}
\renewcommand{\arraystretch}{1.1}
\caption{Ablation study on rethinking strategy for termination decision.}
\vspace{-2mm}
\resizebox{.95\linewidth}{!}{
\begin{tabular}{lcccc}
\toprule
Variants & TL & NE$\downarrow$ & SR$\uparrow$ & SPL$\uparrow$ \\
\midrule
1) \(w/o\) rethink        & 13.21 & 2.76 & 76.40 & 65.73 \\
\multicolumn{1}{>{\cellcolor{mycolor}}l}{2) \(w/\) rethink}     
& \multicolumn{1}{>{\cellcolor{mycolor}}l}{13.28} 
& \multicolumn{1}{>{\cellcolor{mycolor}}l}{2.74} 
& \multicolumn{1}{>{\cellcolor{mycolor}}l}{76.81} 
& \multicolumn{1}{>{\cellcolor{mycolor}}l}{65.85} \\
\bottomrule
\end{tabular}}
\label{tab:rethink-ter}
\vspace{-2mm}
\end{table}

\subsection{Ablation study}
\label{auantitative analysis}
To investigate the effect of each component of VLN-AIEP, we conduct various ablation experiments and report the results on the R2R val-unseen split. Unless otherwise specified, the rethinking strategy for termination is disabled so that progress-aware matching and trajectory representation can be examined independently.

\vspace {1mm}
\noindent{\bf Effect of Progress-Aware fine-grained Matching (PAM).} To evaluate PAM, we test three trajectory representations with and without PAM. As shown in Table~\ref{tab:overall}, comparisons between rows 1\&4, 2\&5, and 3\&6 show that PAM is effective across trajectory representations. Without extra progress supervision, PAM suppresses executed content and emphasizes pending content, leading to better navigation performance.

\vspace {1mm}
\noindent{\bf Effect of landmark-anchored trajectory representation.} 
Rows 5 and 6 of Table~\ref{tab:overall} isolate landmarks while keeping PAM, panorama features, and orientation-aware features fixed. Landmarks improve SR by 2.38 percentage points and SPL by 1.91 percentage points. This supports their role as spatial references for associating the visited trajectory with executed instruction content.

\begin{figure*}[t]
  \centering
   \includegraphics[width=0.9\linewidth]{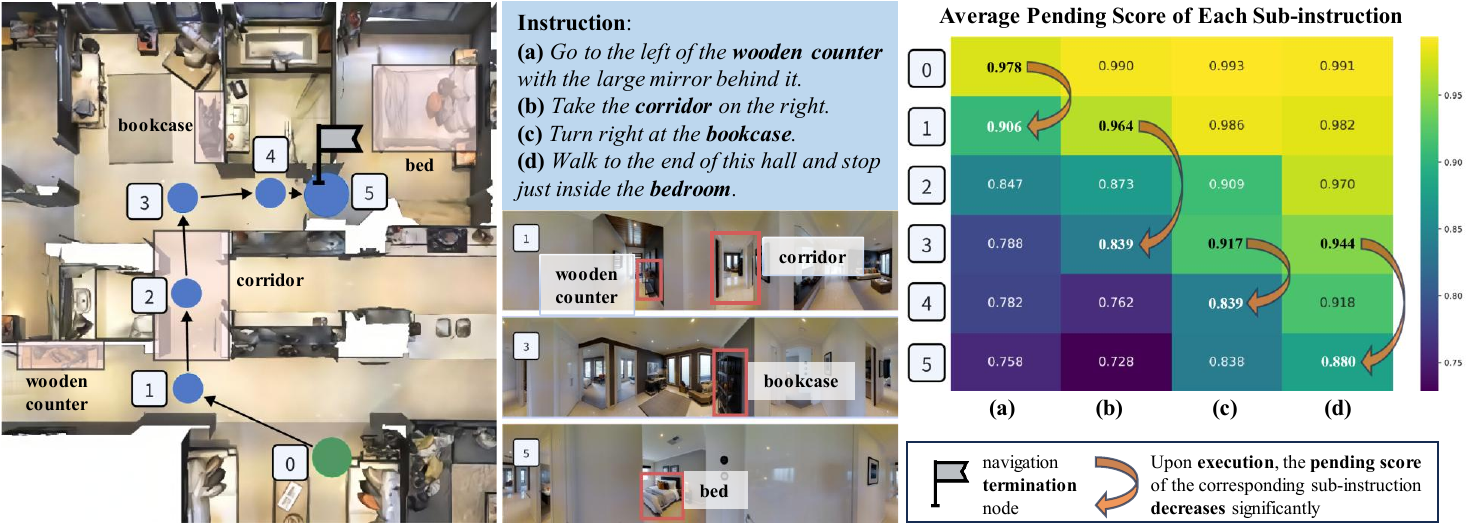}
    \vspace{-1mm}
   \caption{An R2R example of instruction-execution progress estimation. The left panel shows the navigation trajectory. The middle panel shows observations at key steps 1, 3, and 5. Landmark labels identify spatial references mentioned in the instruction. The right panel shows the average pending score for each of four sub-instructions. Each score decreases after the corresponding sub-instruction is executed. This indicates that VLN-AIEP accurately perceives instruction-execution progress. For visualization, the four sentences are treated as four sub-instructions. The model does not use this division during training.}
   \label{fig-v}
   \vspace{-3mm}
\end{figure*}

 \begin{figure}[t]
  \centering
   \includegraphics[width=0.98\linewidth]{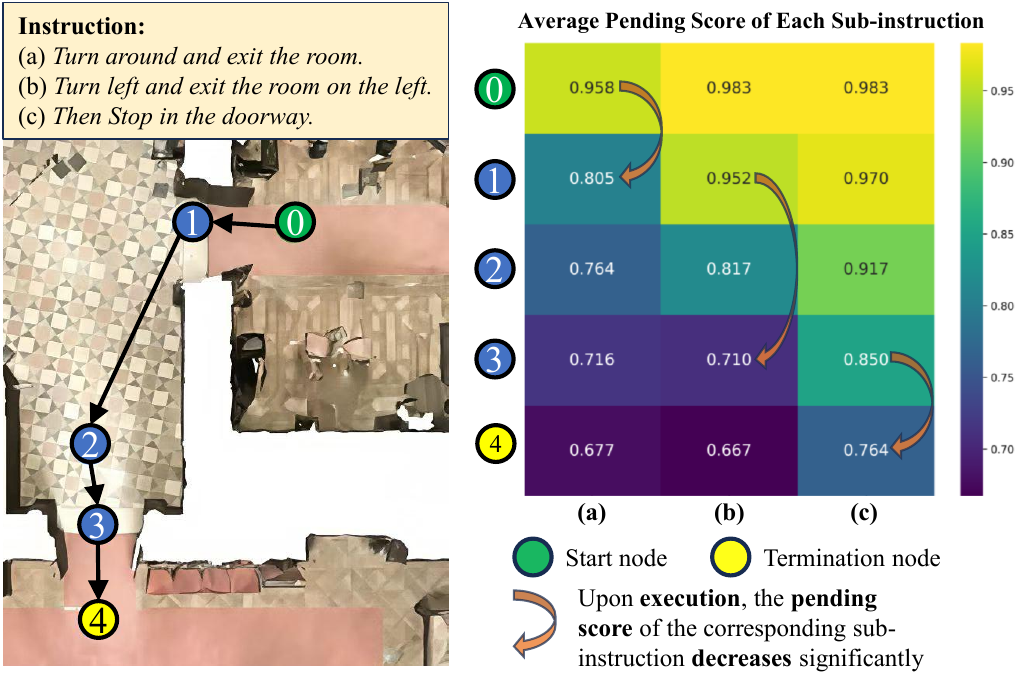}
    \vspace{-1mm}
   \caption{An R2R example of instruction-execution progress estimation. The left panel shows the navigation trajectory, with the instruction divided into three sub-instructions for visualization. The right panel shows their average pending scores. Each score decreases after the corresponding sub-instruction is executed, indicating accurate execution progress estimation.}
   \label{fig-v2}
   \vspace{-2mm}
\end{figure}

 \begin{figure}[t]
  \centering
   \includegraphics[width=0.9\linewidth]{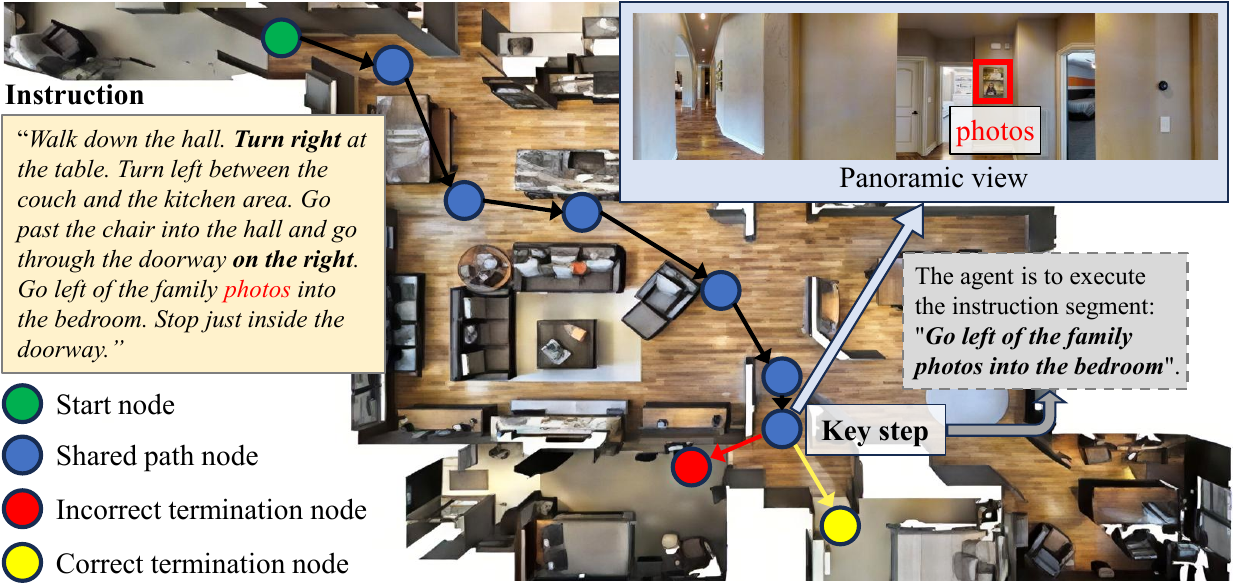}
   \vspace{-1mm}
   \caption{Navigation trajectories of PRET~\cite{lu2024pret} and VLN-AIEP under the same instruction. At the key step, the pending instruction requires a left turn beside the \textit{photos} landmark. PRET remains influenced by the executed phrase, turns right again, and stops incorrectly at the red node. VLN-AIEP uses \textit{photos} as a spatial reference, focuses on the pending instruction, turns left into the target bedroom, and terminates correctly at the yellow node.}
   \label{fig-case2}
   \vspace{-2mm}
\end{figure}

\begin{figure}
  \centering
   \includegraphics[width=0.9\linewidth]{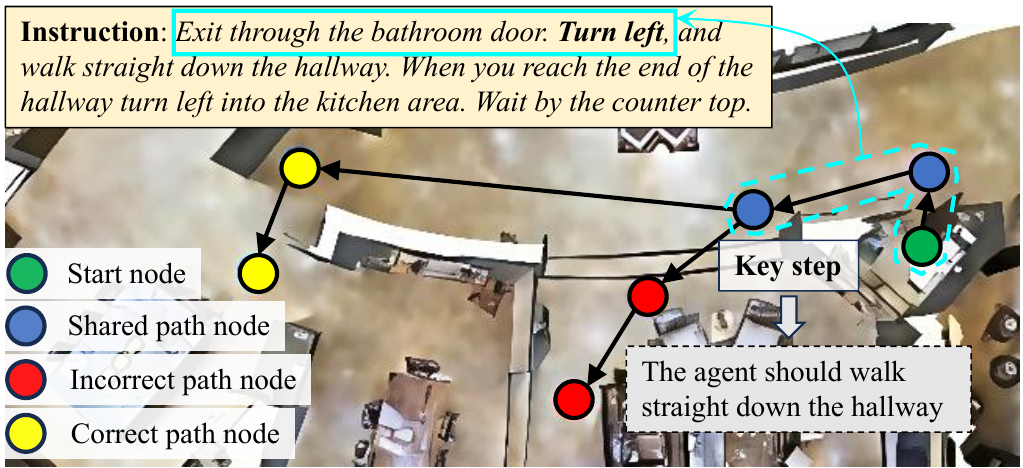}
    \vspace{-1mm}
   \caption{Navigation trajectories of PRET~\cite{lu2024pret} and VLN-AIEP under the same instruction. At the key step, both agents have completed the required left turn and should \textit{walk straight down the hallway}. PRET remains influenced by the executed instruction and turns left again, resulting in the incorrect trajectory shown in red. VLN-AIEP focuses on the pending instruction and walks down the hallway, resulting in the correct trajectory shown in yellow.}
   \label{fig-case3}
\end{figure}

\begin{table}[t]
\centering
\setlength{\tabcolsep}{3pt}  
\renewcommand{\arraystretch}{1.1}
\begin{minipage}{1\linewidth}
    \centering
    \setlength{\tabcolsep}{10pt}
    \renewcommand{\arraystretch}{1.1}
    \vspace{-1mm}
    \caption{SR on val-unseen split of R2R~\cite{anderson2018vision} with and without PAM under perturbation.}
    \vspace{-2mm}
    \resizebox{0.95\columnwidth}{!}{
    \begin{tabular}{cccc}
    \toprule
    w/o PAM & w/o PAM (perturbed) & w/ PAM & w/ PAM (perturbed) \\
    \midrule
    74.2 & 72.3 \textcolor{red}{(-1.9)} & 76.4 & 75.5 \textcolor{red}{(-0.9)} \\
    \bottomrule
    \end{tabular}}
    \label{tab:pam_sr_perturb}
\end{minipage}
\vspace{-4mm}
\end{table}

\begin{figure*}[t]
  \centering
  \includegraphics[width=0.65\linewidth]{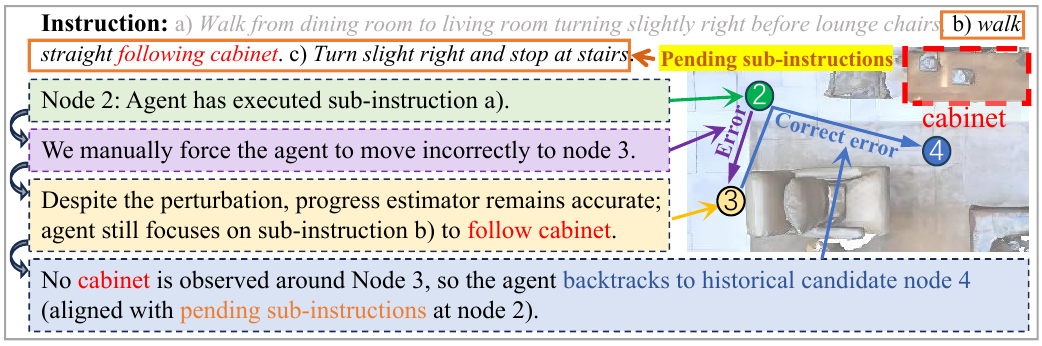}
  \vspace{-2mm}
  \caption{A representative case showing how PAM supports recovery from a manually induced erroneous deviation. After the agent is forced to move from node 2 to node 3, the agent still focuses on the pending sub-instruction ``walk straight following cabinet''. Since no cabinet is observed around node 3, the agent backtracks to historical candidate node 4, which is aligned with the pending instruction at node 2. Notably, after backtracking, the historical trajectory is corrected to the detour-free path 2$\rightarrow$4, rather than 2$\rightarrow$3$\rightarrow$2$\rightarrow$4, for subsequent decision-making.}
  \label{fig-8}
  \vspace{-3mm}
\end{figure*}

\begin{figure}[t]
  \centering
  \includegraphics[width=0.98\linewidth]{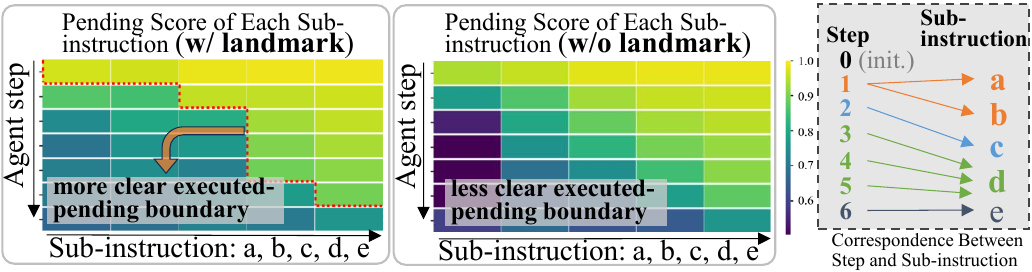}
  \vspace{-2mm}
  \caption{A representative visualization example showing the effect of landmarks as spatial references for progress estimation. Compared with the setting without landmarks, incorporating landmarks produces a clearer boundary between executed and pending sub-instructions, indicating more accurate progress perception.
  }
  \label{fig9}
  \vspace{-2mm}
\end{figure}

\vspace {1mm}
\noindent{\bf Effect of combining I2T and T2I attention.} Table~\ref{tab:attention} shows that removing I2T or T2I reduces SR to 72.62\% or 73.81\%, respectively. Using both branches achieves 76.40\% SR and improves NE and SPL over either single branch. This supports their complementary roles in suppressing instruction content that matches the visited trajectory.

\vspace {1mm}
\noindent{\bf Effect of landmark-anchored MLM. }
Table~\ref{tab:MLM} shows that landmark-anchored MLM improves SR by 4.08 percentage points and SPL by 1.96 percentage points. This supports its role in strengthening the association between landmark tokens and visual trajectory features.

\vspace {1mm}
\noindent{\bf Effect of the rethinking strategy for termination decision. }
Table~\ref{tab:rethink-ter} shows that rethinking improves SR by 0.41 percentage points. These modest gains indicate that rethinking is an auxiliary strategy for refining the final stopping decision.
Additional ablation results are provided in Appendix~B of the supplementary material.

 \subsection{Visualization}
 \label{vis}

Figures~\ref{fig-v} and~\ref{fig-v2} present visualizations of instruction-execution progress estimation in two R2R examples. In Figure~\ref{fig-v}, instruction landmarks provide spatial references that associate the visited trajectory with executed content. Across both examples, the pending score of each sub-instruction decreases after its execution, indicating that VLN-AIEP accurately estimates instruction-execution progress.

Figures~\ref{fig-case2} and~\ref{fig-case3} compare VLN-AIEP with PRET~\cite{lu2024pret} under the same instructions. In Figure~\ref{fig-case2}, the agent must turn left beside the ``photos'' landmark. PRET remains influenced by the executed ``turn right'' segment, turns right again, and stops incorrectly. VLN-AIEP uses ``photos'' as a spatial reference, focuses on the pending content, and enters the target bedroom. In Figure~\ref{fig-case3}, both agents have completed the required left turn and should proceed straight. PRET repeats the left turn, producing the incorrect trajectory shown in red, whereas VLN-AIEP proceeds straight along the correct trajectory shown in yellow. These examples demonstrate that VLN-AIEP suppresses interference from executed content and focuses on pending instructions. Additional visualization results are provided in Appendix~D of the supplementary material.

\subsection{Discussion}

\noindent 
\textbf{Resilience to action perturbations.}
We assess PAM's resilience to random perturbations, where the agent is forced to take incorrect actions. As shown in Table~\ref{tab:pam_sr_perturb}, our PAM effectively mitigates performance degradation caused by erroneous decisions. Figure~\ref{fig-8} further visualizes how PAM facilitates error correction under perturbation. Notably, after a wrong transition, PAM backtracks to the correct historical node and reconstructs a detour-free trajectory by discarding the erroneous step, ensuring that subsequent decision-making is conditioned on the correct navigation history rather than the corrupted path.

\vspace {1mm}
\noindent 
\textbf{Influence of landmarks.} We further investigate the role of landmarks in progress estimation. As illustrated in Figure~\ref{fig9}, incorporating landmarks produces a clearer boundary in the average pending scores between executed and pending sub-instructions, while removing landmarks blurs this boundary. Consequently, the average score gradient at the executed-pending boundary decreases from 0.1101 to 0.0786. This indicates that landmarks serve as effective semantic-spatial references for associating historical movements with executed instruction parts, thereby improving progress awareness.

\section{Conclusion}

In this paper, we proposed VLN-AIEP, which couples instruction-execution progress estimation with navigation decision-making. It suppresses executed instruction content and matches candidate locations with pending content. This coupling enables progress estimation to be learned directly from navigation supervision without explicit progress annotations. A landmark-anchored trajectory representation further supports progress estimation using landmark semantics and geometry. Experiments on R2R, RxR, CVDN, and R2R-CE demonstrate the effectiveness of VLN-AIEP.
\bibliographystyle{IEEEtran}
\bibliography{VLN}

\vfill

\end{document}